\documentclass[10pt,twocolumn]{article}

\usepackage{graphicx}
\usepackage{multirow}
\usepackage{amsmath,amssymb}
\usepackage{booktabs}
\usepackage{array}
\usepackage{tabularx}
\usepackage{xcolor}
\usepackage{listings}
\usepackage{microtype}
\usepackage[numbers,sort&compress]{natbib}
\usepackage[colorlinks=true,citecolor=blue,linkcolor=blue,urlcolor=blue]{hyperref}
\usepackage{url}

\graphicspath{{figures/}}

\providecommand{\orcidlink}[1]{\href{https://orcid.org/#1}{#1}}

\lstdefinelanguage{SQL}{
  morekeywords={SELECT,FROM,WHERE,AND,OR,IN,NOT,EXISTS,GROUP,BY,ORDER,HAVING,ASC,DESC,LIMIT,JOIN,ON,AS,SUM},
  sensitive=false,
  morecomment=[l]{--},
  morestring=[b]',
}

\newcolumntype{L}{>{\raggedright\arraybackslash}X}

\title{\textbf{TiCard: Deployable EXPLAIN-only Residual Learning for Cardinality Estimation}}

\author{
Qizhi Wang (\orcidlink{0009-0004-1346-5066})\\
PingCAP, Data \& AI-Innovation Lab, Beijing, China\\
\texttt{qizhi.wang@pingcap.com}
}

\date{}

\begin{document}
\maketitle

\begin{abstract}
Cardinality estimation is a key bottleneck for cost-based query optimization, yet deployable improvements remain difficult: classical estimators miss correlations, while learned estimators often require workload-specific training pipelines and invasive integration into the optimizer.
This paper presents \textbf{TiCard}, a low-intrusion, correction-based framework that augments (rather than replaces) a database's native estimator.
TiCard learns multiplicative residual corrections using \texttt{EXPLAIN}-only features, and uses \texttt{EXPLAIN ANALYZE} only for offline labels.
We study two practical instantiations: (i) a Gradient Boosting Regressor for sub-millisecond inference, and (ii) TabPFN, an in-context tabular foundation model that adapts by refreshing a small reference set without gradient retraining.
On TiDB with TPC-H and the Join Order Benchmark, in a low-trace setting (263 executions total; 157 used for learning), TiCard improves operator-level tail accuracy substantially: P90 Q-error drops from 312.85 (native) to 13.69 (TiCard-GBR), and P99 drops from 37,974.37 to 3,416.50 (TiCard-TabPFN), while a join-only policy preserves near-perfect median behavior.
We position TiCard as an AI4DB building block focused on deployability: explicit scope, conservative integration policies, and an integration roadmap from offline correction to in-optimizer use.
\end{abstract}


\section{Introduction}
\label{sec:introduction}

Cardinality estimation (CE)---predicting the number of rows produced by each operator---is central to cost-based query optimization, affecting join ordering, physical operator choice, and memory management~\cite{selinger1979,leis2015}.
Despite decades of work, CE remains brittle in modern analytical workloads, primarily because independence assumptions and limited statistics struggle with multi-column predicates and cross-table correlations~\cite{ioannidis2003,leis2018job}.

From an \emph{AI4DB} perspective, the challenge is not only improving accuracy but doing so in a way that is \emph{deployable}:
learned estimators can be accurate, yet are often costly to train, sensitive to workload drift, and require deep integration into the optimizer's enumeration loop~\cite{kipf2019learned,yang2020neurocard,marcus2021bao}.
In practice, database teams frequently prefer incremental, low-risk changes that preserve existing optimizer behavior and can be rolled out conservatively.

This paper proposes a pragmatic framing: treat the native optimizer as a strong prior and learn only its \emph{residual error}.
We introduce \textbf{TiCard}, a correction-based CE framework that learns multiplicative adjustments on top of the optimizer estimate.
Crucially, TiCard's feature pipeline is derived from \texttt{EXPLAIN} only, enabling a low-intrusion path that leverages existing database interfaces.
\texttt{EXPLAIN ANALYZE} is used solely for offline label collection.

\subsection{Scope and deployability goals}
\label{sec:scope}

We explicitly scope this work to the setting that is most actionable for deployment teams:
\begin{itemize}
\item \textbf{Low intrusion:} learn from existing interfaces (\texttt{EXPLAIN} / \texttt{EXPLAIN ANALYZE}) without requiring a new optimizer or deep runtime instrumentation.
\item \textbf{Data efficiency:} operate under a low-trace regime where executed-query labels are expensive (hundreds of executions, not thousands).
\item \textbf{Safety controls:} support conservative policies (e.g., join-only correction, blending with fallback) to preserve strong baseline behavior.
\item \textbf{Evaluation focus:} we report \emph{offline}, operator-level CE accuracy on collected plans; we do not claim end-to-end plan-quality or latency gains without full integration into the optimizer.
\end{itemize}

This scope is not a limitation to hide; it is a design choice motivated by deployability.
We therefore also provide an integration roadmap that describes how to use TiCard-style corrections inside a live optimizer, and where the engineering risks and overheads arise.

\subsection{Contributions}
\label{sec:contributions}

Our main contributions are:
\begin{enumerate}
\item \textbf{EXPLAIN-only correction formulation:} We frame CE as learning multiplicative residual corrections using a leakage-free feature pipeline derived from \texttt{EXPLAIN} plans.
\item \textbf{Deployable model choices:} We study two complementary instantiations---TabPFN in-context learning (fast refresh without gradient retraining) and Gradient Boosting Regression (very fast inference).
\item \textbf{Conservative integration policies:} We define and evaluate practical policies (join-only correction, blending, and a two-stage design for zero-cardinality cases) aimed at controlling regressions.
\item \textbf{Empirical evaluation in a low-trace regime:} On TiDB with TPC-H and JOB, we show large tail improvements at the operator level using only 157 training executions, and quantify setup/training and inference costs.
\item \textbf{Integration roadmap:} We outline a path from offline correction to online use, with overhead and risk considerations that matter for real deployments.
\end{enumerate}

\section{Background}
\label{sec:background}

\subsection{Cardinality estimation and Q-error}
\label{sec:qerror}

CE quality is commonly measured by \textbf{Q-error}~\cite{moerkotte2009}:
\begin{equation}
\text{Q-error} = \max\left(\frac{\text{actual}}{\text{estimated}}, \frac{\text{estimated}}{\text{actual}}\right).
\end{equation}
Q-error is scale-invariant and symmetric for over/underestimation.
We apply a standard guard \(\max(\cdot,1)\) to keep Q-error well-defined when the true or predicted cardinality is zero.

\subsection{Plan interfaces: \texttt{EXPLAIN} and \texttt{EXPLAIN ANALYZE}}
\label{sec:explain}

Many systems expose estimated and actual operator cardinalities through plan interfaces.
In TiDB, \texttt{EXPLAIN} returns the chosen plan with estimated rows (\texttt{estRows}), and \texttt{EXPLAIN ANALYZE} executes the query and reports actual rows (\texttt{actRows}).
TiCard uses \texttt{EXPLAIN} to construct features and \texttt{EXPLAIN ANALYZE} only to obtain labels.

\section{TiCard: Correction-based CE with EXPLAIN-only Features}
\label{sec:ticard}

\begin{figure}[t]
\centering
\includegraphics[width=\linewidth]{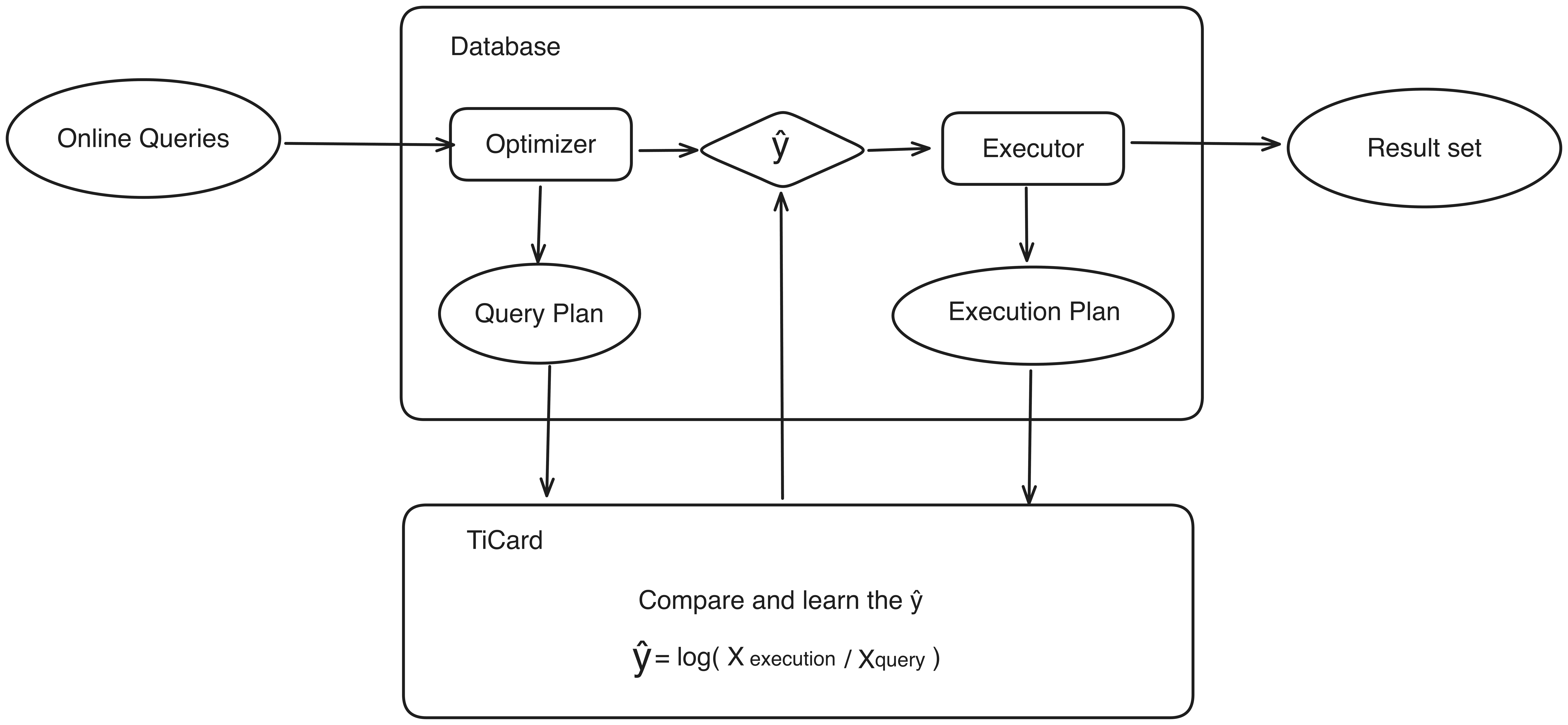}
\caption{TiCard pipeline: plan extraction, EXPLAIN-only feature engineering, correction-target construction, and model setup/training and inference.}
\label{fig:architecture}
\end{figure}

\subsection{Overview}

Given an operator node \(q\) in a query plan, let \(E(q)\) be the native optimizer estimate from \texttt{EXPLAIN} and \(C(q)\) be the true cardinality observed from \texttt{EXPLAIN ANALYZE}.
TiCard learns a function \(f(\cdot)\) that predicts a log-space correction target:
\begin{equation}
\hat{y}(q) \approx \log\left(\frac{1+C(q)}{1+E(q)}\right).
\label{eq:target}
\end{equation}
At inference time, the corrected estimate is:
\begin{equation}
\widehat{C}(q) = (1+E(q))\cdot \exp(\hat{y}(q)) - 1.
\label{eq:corrected}
\end{equation}
This residual formulation leverages the native estimator as a prior and focuses learning capacity on systematic optimizer errors.

\subsection{Feature engineering (leakage-free)}
\label{sec:features}

All model inputs are derived from \texttt{EXPLAIN} only.
We extract per-operator features that capture local properties and global plan context, including:
\begin{itemize}
\item \textbf{Native estimate:} optimizer estimated output cardinality (and its log transform).
\item \textbf{Plan structure:} depth/position features and ratios relative to total estimated rows.
\item \textbf{Operator attributes:} operator type (scan/filter/join/aggregation), table name, task type, and join/scan sub-types when applicable.
\end{itemize}

We split datasets by \textbf{query execution} (not by operator) to avoid leakage: all operator nodes from a query belong to the same split.
For preprocessing, we use one-hot encoding for categorical features, standard scaling for numerical features, and simple missing-value handling (zero or \texttt{unknown}).
To reduce overfitting in the low-trace regime, we apply feature selection (\texttt{SelectKBest} with $k$ tuned on a validation set).
For correction learning, we optionally stabilize the training target in Eq.~\ref{eq:target} with a robust IQR-based clipping range; evaluation still reports on all test samples without filtering.

\subsection{Model instantiations}
\label{sec:models}

\textbf{TiCard-GBR} uses gradient boosting regression~\cite{friedman2001} for maximum inference speed.
This choice targets production-grade overhead constraints, where CE must be computed frequently.

\textbf{TiCard-TabPFN} uses TabPFN~\cite{hollmann2023tabpfn}, a transformer pre-trained for tabular prediction, in an in-context learning mode.
TabPFN adapts by refreshing the reference set (the \emph{setup} data) without gradient retraining, which is attractive under workload drift when labels are available but retraining pipelines are costly.

\paragraph{TabPFN regression setup.}
Although TabPFN is often discussed in the context of classification, we use a regression variant (\texttt{TabPFNRegressor}) to predict continuous targets.
Specifically, the target is the continuous log-space correction in Eq.~\ref{eq:target} (or \(\log(1+\text{act})\) for the direct ablation).
We use a small ensemble (\texttt{n\_estimators=8}) for throughput and run inference without fine-tuning.

\subsubsection{Terminology: ``setup'' vs.\ ``training''}
\label{sec:terminology}

To avoid ambiguity, we distinguish two learning paradigms:
\begin{itemize}
\item \textbf{Setup / in-context learning (TabPFN):} \texttt{fit()} stores the reference set for attention-based inference; no gradient descent or parameter updates occur.
\item \textbf{Training (GBR and neural baselines):} model parameters are learned via standard optimization (boosting or backpropagation).
\end{itemize}
Throughout the paper, ``training executions'' refer to the same labeled traces, but they are consumed as setup data for TabPFN and as training data for GBR/neural models.

\subsection{Conservative integration policies}
\label{sec:policies}

Correction is a \emph{design space} rather than a single rule.
To improve deployability, TiCard supports conservative policies that can be selected and tuned on a validation set:
\begin{itemize}
\item \textbf{Join-only correction:} apply corrections only to join operators and fall back to native estimates for other nodes, preserving strong median behavior.
\item \textbf{Blending:} combine corrected and native estimates with a bounded rule (e.g., clamp correction factor within a range) to reduce risk from outliers.
\item \textbf{Two-stage zero handling:} a classifier for zero vs.\ non-zero cardinality followed by regression, reducing instability around zero-cardinality operators.
\end{itemize}

\section{Experimental Evaluation (Offline Operator-level)}
\label{sec:evaluation}

\subsection{Setup}
\label{sec:setup}

\textbf{System:} TiDB v8.5.2.
After loading datasets we run \texttt{ANALYZE TABLE} to populate statistics.

\textbf{Benchmarks:} TPC-H (scale factor 1) and JOB~\cite{tpch,leis2018job}.
For TPC-H we execute 22 query templates with multiple parameterizations (150 executions total); for JOB we execute 113 distinct queries (113 executions).
Overall, we collect 263 query executions and 6114 operator samples.

\textbf{Split:} 157 executions for learning (3556 operator samples), 53 for validation (1369), and 53 for test (1189).

\textbf{Baselines:} native TiDB estimates, simple calibration baselines (group scaling and isotonic regression), a LiteCard-style local correction baseline~\cite{yi2025biggerbreadboxefficientcardinality}, and a MADE (NeuroCard-architecture) correction baseline~\cite{yang2020neurocard,germain2015made} trained on our EXPLAIN-derived feature vectors (not NeuroCard's canonical data-driven pipeline).
\emph{Note on LiteCard:} our baseline is an offline analogue of LiteCard's hierarchical local-correction idea (pattern-keyed regressors with fallback). It does not reproduce LiteCard's full online learning loop or planner hooks; we use it to contextualize lightweight correction behavior under the same EXPLAIN-only, offline setting.

\textbf{Metrics:} Q-error statistics (median/mean/P90/P99), quality-band distribution, and wall-clock setup/training and inference time.

\subsection{Overall results}
\label{sec:overall}

\begin{table*}[t]
\centering
\caption{Q-error on the test set (1189 operator samples; no filtering).}
\label{tab:qerror}
\scriptsize
\setlength{\tabcolsep}{3.2pt}
\renewcommand{\arraystretch}{1.05}
\begin{tabular*}{\textwidth}{@{\extracolsep{\fill}}p{0.42\textwidth}rrrr}
\toprule
\textbf{Model} & \textbf{P90} & \textbf{P99} & \textbf{Median} & \textbf{Mean} \\
\midrule
TiDB Default & 312.85 & 37,974.37 & \textbf{1.0030} & 3,045.79 \\
Scale (group) & 226.79 & 16,214.41 & 1.0100 & 1,728.37 \\
Isotonic (group) & 75.70 & 2,706.81 & 1.8247 & 131.20 \\
LiteCard (corr.) & 55.29 & \textbf{2,355.90} & 1.8751 & 140.35 \\
TiCard-TabPFN (our) & 13.82 & 3,416.50 & 1.0406 & 131.90 \\
TiCard-GBR (our) & \textbf{13.69} & 3,812.02 & 1.3158 & \textbf{122.03} \\
TiCard-GBR (join-only, our) & 70.54 & 4,576.90 & 1.0078 & 1,559.91 \\
MADE (NeuroCard arch., corr.) & 202.60 & 12,730.89 & 1.8363 & 988.50 \\
\bottomrule
\end{tabular*}
\end{table*}

TiCard substantially improves tail accuracy at the operator level, while preserving strong baseline behavior via conservative policies.
In particular, TiCard-GBR reduces P90 Q-error from 312.85 to 13.69, and TiCard-TabPFN reduces P99 from 37,974.37 to 3,416.50.
The join-only policy illustrates a deployability trade-off: it preserves TiDB's near-perfect median (1.0078) while still reducing tail errors, but gives up some of the full-correction improvements.

\begin{figure}[t]
\centering
\includegraphics[width=\linewidth]{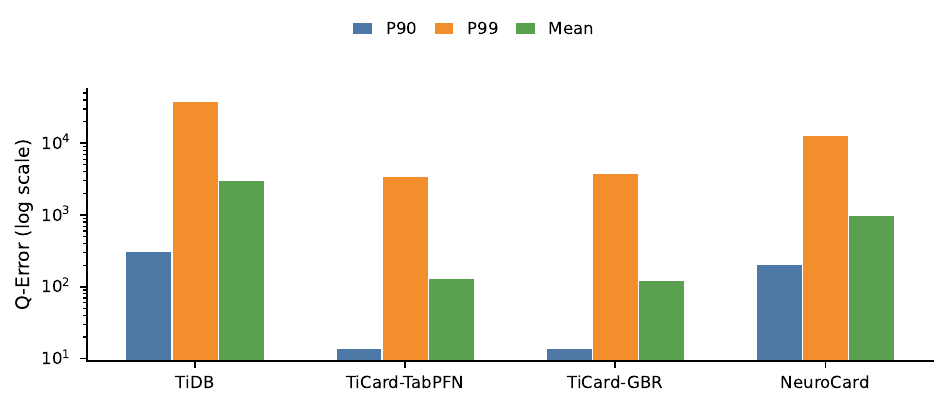}
\caption{Q-error comparison across models (P90, P99, Mean; log scale).}
\label{fig:qerror-comparison}
\end{figure}

\subsection{Distribution analysis}
\label{sec:distribution}

\begin{table*}[t]
\centering
\caption{Q-error distribution on the test set (1189 operator samples).}
\label{tab:distribution}
\scriptsize
\setlength{\tabcolsep}{3.0pt}
\renewcommand{\arraystretch}{1.05}
\begin{tabular*}{\textwidth}{@{\extracolsep{\fill}}p{0.40\textwidth}ccccc}
\toprule
\textbf{Model} & \textbf{Excellent} & \textbf{Good} & \textbf{Fair} & \textbf{Poor} & \textbf{Terrible} \\
& ($\leq$2) & (2--5] & (5--10] & (10--100] & ($>$100) \\
\midrule
TiDB Default & 67.9\% & 5.8\% & 2.6\% & 11.2\% & 12.5\% \\
Scale (group) & 64.3\% & 10.2\% & 4.0\% & 9.2\% & 12.4\% \\
Isotonic (group) & 52.6\% & 18.1\% & 9.1\% & 12.4\% & 7.9\% \\
TiCard-TabPFN (our) & \textbf{78.2\%} & 6.4\% & 3.7\% & 7.4\% & \textbf{4.3\%} \\
TiCard-GBR (our) & 72.8\% & 11.3\% & 4.1\% & 7.3\% & 4.5\% \\
LiteCard (corr.) & 52.5\% & 20.9\% & 6.1\% & 12.4\% & 8.0\% \\
MADE (NeuroCard arch., corr.) & 52.4\% & 23.4\% & 3.4\% & 9.0\% & 11.8\% \\
\bottomrule
\end{tabular*}
\end{table*}

\begin{figure}[t]
\centering
\includegraphics[width=\linewidth]{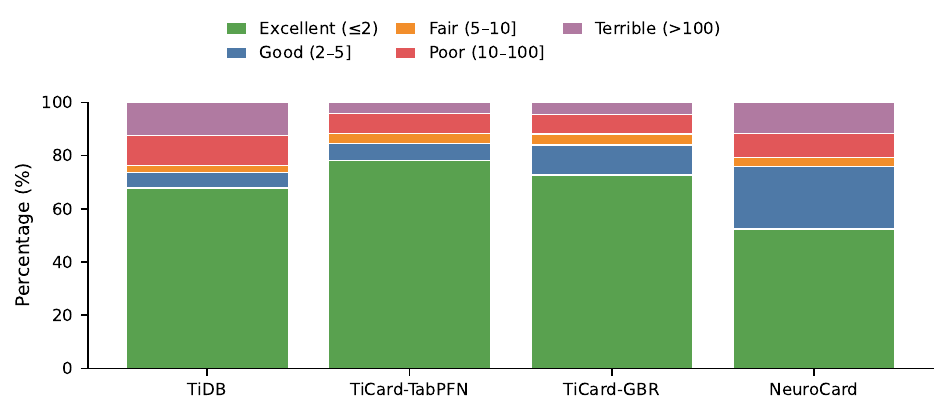}
\caption{Q-error quality bands (percentage of estimates in each range).}
\label{fig:qerror-dist}
\end{figure}

\begin{figure}[t]
\centering
\includegraphics[width=\linewidth]{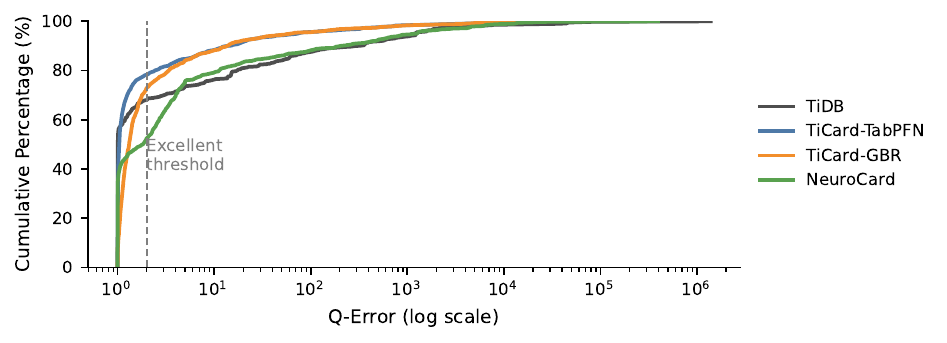}
\caption{Cumulative distribution function (CDF) of Q-error (log scale).}
\label{fig:qerror-cdf}
\end{figure}

TiCard reduces catastrophic errors (Q-error \(>\) 100) from 12.5\% to about 4--5\% and increases the share of excellent estimates (\(\leq 2\)).
This matters for deployability because optimizer failures are often driven by tail events rather than median behavior.

\subsection{Robustness and ablations (no new executions)}
\label{sec:robustness}

Review feedback often asks whether improvements rely on memorization via instance-specific categorical features, how corrections behave across operator types, and whether per-operator corrections translate into plan-level improvements.
We address these with additional analyses computed on the same cached \texttt{EXPLAIN} / \texttt{EXPLAIN ANALYZE} traces (no new query executions).

\subsubsection{Feature inventory and selection}

Table~\ref{tab:feature-inventory} enumerates categorical feature cardinalities in our TiDB plan traces and the resulting feature dimensionality.
After one-hot encoding, the learning pipeline uses a compact feature matrix (14 numeric features after dropping IDs/labels).
Validation-tuned \texttt{SelectKBest} retains $k=10$ features for correction mode; the selected set is dominated by native-estimate and plan-structure signals, consistent with the deployability goal of minimizing feature dependence.

\begin{table*}[t]
\centering
\caption{Feature inventory and selection (offline traces).}
\label{tab:feature-inventory}
\small
\setlength{\tabcolsep}{4.0pt}
\begin{tabular*}{0.72\textwidth}{@{\extracolsep{\fill}}lrr}
\toprule
\textbf{Item} & \textbf{Count} & \textbf{Notes} \\
\midrule
Operator types & 18 & \texttt{operator\_type} categories \\
Task types & 2 & \texttt{task\_type} categories \\
Join types & 5 & \texttt{join\_type} categories \\
Table identities & 4 & extracted when available; often coarse \\
Feature dim (post-encode) & 14 & numeric features used by models \\
Selected $k$ & 10 & chosen on validation set \\
\bottomrule
\end{tabular*}
\end{table*}

For the default split, the 10 selected features are:
\texttt{optimizer\_est\_out}, \texttt{log\_est\_rows}, \texttt{plan\_depth}, \texttt{node\_position}, \texttt{relative\_position}, \texttt{est\_to\_total\_ratio}, and join/scan indicators (\texttt{is\_join}, \texttt{is\_scan}, \texttt{is\_table\_scan}, \texttt{is\_hash\_join}).

\subsubsection{Per-operator breakdown}

Table~\ref{tab:op-breakdown} breaks down tail errors by coarse operator group.
The largest reductions occur on joins (where correlation effects dominate), while scans/aggregations also improve substantially.
This addresses the concern that improvements are confined to a narrow operator class.

\begin{table*}[t]
\centering
\caption{Per-operator breakdown (test split; P90/P99 Q-error).}
\label{tab:op-breakdown}
\scriptsize
\setlength{\tabcolsep}{3.2pt}
\begin{tabular*}{\textwidth}{@{\extracolsep{\fill}}lrrrr}
\toprule
\textbf{Group} & \textbf{P90 (TiDB)} & \textbf{P99 (TiDB)} & \textbf{P90 (TiCard-GBR)} & \textbf{P99 (TiCard-GBR)} \\
\midrule
Join & 1815.32 & 45,568.47 & 59.02 & 6,263.02 \\
Scan & 18.97 & 1,710.98 & 5.90 & 599.68 \\
Filter & 109.51 & 2,937.37 & 21.14 & 1,367.13 \\
Aggregation & 363.18 & 3,514.45 & 7.01 & 307.34 \\
Other & 902.32 & 219,959.70 & 23.01 & 12,863.68 \\
\bottomrule
\end{tabular*}
\end{table*}

\subsubsection{Plan-level proxy: root-node cardinality}

Although we do not integrate into TiDB's plan enumeration loop, we can report a plan-level proxy using the root operator's output cardinality (one root node per query execution).
Table~\ref{tab:root-qerror} shows that independent per-operator correction nevertheless improves root-node tail errors dramatically, suggesting reduced error propagation at the plan output level.

\begin{table}[t]
\centering
\caption{Root-node (plan output) Q-error across queries (53 executions in the test split).}
\label{tab:root-qerror}
\small
\setlength{\tabcolsep}{4.0pt}
\begin{tabular}{@{}lrrrr@{}}
\toprule
\textbf{Model} & \textbf{Median} & \textbf{P90} & \textbf{P99} & \textbf{Mean} \\
\midrule
TiDB Default & 1.00 & 55.46 & 989.68 & 62.29 \\
TiCard-GBR & 1.26 & 2.38 & 23.11 & 2.79 \\
\bottomrule
\end{tabular}
\end{table}

\subsubsection{Query-level worst-case (plan-aware) proxies}
\label{sec:plan-aware}

To further align evaluation with plan outcomes, we report query-level worst-case errors, which are more predictive of optimizer ``disasters'' than operator-marginal averages.
For each query execution, we compute (i) the maximum operator Q-error in the plan and (ii) the maximum join-node Q-error (since joins dominate plan sensitivity).
Table~\ref{tab:query-worst} summarizes the distribution over the 53 test executions.
These metrics are not a substitute for executing alternative plans, but they are plan-aware in the sense that they aggregate errors \emph{within} a plan and emphasize worst-case nodes.

\begin{table*}[t]
\centering
\caption{Query-level worst-case proxies over the 53 test executions.}
\label{tab:query-worst}
\small
\setlength{\tabcolsep}{3.0pt}
\begin{tabular*}{\textwidth}{@{\extracolsep{\fill}}p{0.28\textwidth}p{0.22\textwidth}rrrr}
\toprule
\textbf{Metric} & \textbf{Model} & \textbf{Median} & \textbf{P90} & \textbf{P99} & \textbf{Mean} \\
\midrule
\multirow{2}{*}{Max operator Q-error} & TiDB Default & 90.93 & 4,560.41 & 710,991.63 & 29,826.90 \\
 & TiCard-GBR & 5.49 & 1,668.87 & 16,466.88 & 1,050.21 \\
\multirow{2}{*}{Max join-node Q-error} & TiDB Default & 52.22 & 2,508.54 & 743,000.10 & 30,912.90 \\
 & TiCard-GBR & 2.38 & 843.77 & 13,846.29 & 862.28 \\
\bottomrule
\end{tabular*}
\end{table*}

Plan-aware metrics such as \emph{P-error} can correlate better with end-to-end outcomes in some settings; however, they typically require plan-cost models and/or enumerator access to define and evaluate alternative plans.
Within our EXPLAIN-only scope, we therefore report root-node error (Table~\ref{tab:root-qerror}) and query-level worst-case proxies (Table~\ref{tab:query-worst}) as partial evidence, and defer full planner-in-the-loop metrics to future work.

\subsubsection{TPC-H parameter holdout (template-stratified)}

To address concerns about memorization on TPC-H due to repeated template executions, we run a template-stratified execution holdout: for each of the 22 templates, we hold out a subset of parameterizations as test and train on the remaining parameterizations (no template is entirely absent from training).
On this holdout (96/25/29 executions for train/val/test), TiCard-GBR improves tail errors from P90 23.83 to 1.29 and P99 2,113.06 to 16.75 (median 1.01 to 1.05).
This indicates that the correction signal generalizes across parameters rather than relying on exact query instances.

\subsection{Improvement factors over the native optimizer}
\label{sec:improvement}

\begin{table}[t]
\centering
\caption{Improvement factors over native TiDB (TiDB Q-error / TiCard Q-error).}
\label{tab:improvement}
\setlength{\tabcolsep}{5.0pt}
\begin{tabular}{@{}lcc@{}}
\toprule
\textbf{Metric} & \textbf{TiCard-TabPFN} & \textbf{TiCard-GBR} \\
\midrule
P90 & 22.6$\times$ & 22.9$\times$ \\
P99 & 11.1$\times$ & 10.0$\times$ \\
Mean & 23.1$\times$ & 25.0$\times$ \\
\bottomrule
\end{tabular}
\end{table}

The largest gains occur on the same tail events where classical independence assumptions cause multiplicative error propagation (especially across joins).
From a deployment standpoint, this is desirable: the primary goal is to reduce optimizer ``disasters'' rather than to chase marginal improvements on already-correct estimates.

\subsection{Efficiency: setup/training and inference}
\label{sec:efficiency}

\begin{table*}[t]
\centering
\caption{Setup/training and inference performance (test set inference over 1189 operator nodes).}
\label{tab:efficiency}
\scriptsize
\setlength{\tabcolsep}{3.2pt}
\begin{tabular*}{\textwidth}{@{\extracolsep{\fill}}p{0.42\textwidth}rrr}
\toprule
\textbf{Model} & \textbf{Setup/Training (s)} & \textbf{Inference (s)} & \textbf{Samples/sec} \\
\midrule
TiCard-TabPFN (our) & 2.29 & 2.41 & 493 \\
TiCard-GBR (our) & 0.65 & 0.0022 & 551K \\
LiteCard (corr.) & 0.01 & 0.022 & 52.9K \\
MADE (NeuroCard arch., corr.) & 5.87 & 0.38 & 3.10K \\
\bottomrule
\end{tabular*}
\end{table*}

GBR provides very fast inference suitable for tight optimization loops, while TabPFN offers fast refresh without gradient retraining but at a much higher inference cost.
For a plan with 50 operators, the node-level inference overhead is roughly \(\sim\)0.09ms (GBR) versus \(\sim\)101ms (TabPFN), suggesting different deployment positions: online optimization vs.\ offline analysis or amortized planning.

\subsection{Ablation: direct prediction vs.\ correction}
\label{sec:ablation}

To isolate the effect of \emph{residual learning}, we compare predicting cardinalities directly (\(\log(1+\text{act})\)) with predicting correction targets (Eq.~\ref{eq:target}) under the same features and model family.

\begin{table*}[t]
\centering
\caption{Ablation on the test set (1189 operator samples).}
\label{tab:ablation}
\scriptsize
\setlength{\tabcolsep}{3.0pt}
\begin{tabular*}{\textwidth}{@{\extracolsep{\fill}}p{0.34\textwidth}p{0.14\textwidth}rrrr}
\toprule
\textbf{Model} & \textbf{Mode} & \textbf{P90} & \textbf{P99} & \textbf{Median} & \textbf{Mean} \\
\midrule
TiDB Default & native & 312.85 & 37,974.37 & 1.0030 & 3,045.79 \\
TabPFN & direct & 16.64 & \textbf{2,280.87} & \textbf{1.0379} & 528.34 \\
TiCard-TabPFN (our) & correction & \textbf{13.82} & 3,416.50 & 1.0406 & \textbf{131.90} \\
GBR & direct & 16.82 & \textbf{1,736.23} & \textbf{1.2956} & \textbf{106.84} \\
TiCard-GBR (our) & correction & \textbf{13.69} & 3,812.02 & 1.3158 & 122.03 \\
\bottomrule
\end{tabular*}
\end{table*}

Direct prediction and correction are complementary: correction improves P90 for both models but can worsen the extreme tail (P99) relative to direct learning.
This supports a deployability viewpoint: correction should be paired with conservative policies (e.g., join-only, blending) and validated for regressions.

\paragraph{Zero-handling policy.}
We implement a two-stage option (zero vs.\ non-zero classifier, then regression), but on our default split it does not improve and can slightly worsen P99 due to false positives (predicting zero when the true cardinality is non-zero).
We therefore report results without the two-stage override by default and treat robust zero handling as future work; the primary stabilization used here is the \(\log(1+x)\) target transform in Eq.~\ref{eq:target}.

\subsection{Limitations of the offline study}
\label{sec:limitations}

Our evaluation is intentionally scoped:
\begin{itemize}
\item \textbf{No end-to-end planner impact:} we do not measure join-order changes or query latency improvements, which require integration into TiDB's plan enumeration and costing loop.
\item \textbf{Operator independence:} we predict per-operator corrections independently; future work should model error propagation through the plan tree.
\item \textbf{Out-of-distribution patterns:} novel operators or rare plan motifs can still produce large errors; uncertainty estimation and fallback policies are important for production use.
\end{itemize}

We emphasize that CE accuracy is necessary but not sufficient for end-to-end performance: plan quality also depends on the enumerator search space, cost model, and runtime behavior.
To provide partial evidence without invasive integration, we include a plan-level proxy based on root-node cardinality (Table~\ref{tab:root-qerror}), which improves substantially and suggests reduced propagation at the plan output level~\cite{ioannidis1991}.

\paragraph{Optimizer evolution and model refresh.}
Because TiCard conditions on the native estimate \(E(q)\), changes to the optimizer's estimator or cost model can shift the residual distribution.
In practice this can be handled with lightweight refresh: GBR retrains in seconds on hundreds of executions, and TabPFN refreshes by replacing the in-context reference set.
We recommend feature and model versioning tied to optimizer releases to detect and manage such drift.

\section{Integration Roadmap: From Offline Correction to Online Use}
\label{sec:roadmap}

To make TiCard actionable for ML-for-DB deployments, we outline an integration path that preserves optimizer control and enables staged rollout:
\begin{enumerate}
\item \textbf{Label collection:} select and execute representative queries (or sampled production queries) to collect \texttt{EXPLAIN} / \texttt{EXPLAIN ANALYZE} pairs.
\item \textbf{Model training/refresh:} train GBR periodically (seconds for hundreds of executions), or refresh TabPFN's reference set when drift is detected.
\item \textbf{Inference placement:} cache per-(sub)plan-node features and predictions; apply corrections at the cardinality estimation boundary, not by replacing the optimizer.
\item \textbf{Safety controls:} start with join-only correction and bounded blending; add uncertainty-based fallback once confidence estimates are available.
\item \textbf{Measurement:} evaluate plan stability (join-order changes), compilation overhead, and latency/throughput on shadow traffic before enabling corrections broadly.
\end{enumerate}

This roadmap is consistent with the AI4DB principle that a learned component should be \emph{incremental}, \emph{observable}, and \emph{reversible}.

\subsection{Why we do not report end-to-end optimizer results in this paper}
\label{sec:why-no-e2e}

Reviewers often request end-to-end measurements (chosen plans, plan stability, and query latency).
We agree these are important for deployment, but we intentionally do not include them here because they require a \emph{different} artifact: an in-planner integration into TiDB's enumeration and costing loop.
Concretely, end-to-end evaluation would require (i) injecting corrected cardinalities into the optimizer's cardinality interface for candidate plan enumeration, (ii) ensuring feature extraction and inference are available inside the planning critical path (including caching and concurrency), and (iii) defining safe rollback and gating policies.
These changes are non-trivial engineering work and introduce confounders from the cost model, plan enumeration search space, and runtime effects~\cite{lan2021survey}.
As a result, end-to-end outcomes can be difficult to attribute cleanly to CE correction alone.

Instead, this paper focuses on a deployable \emph{building block} and provides (a) strong operator-level tail reductions, (b) a plan-level proxy via root-node Q-error (Table~\ref{tab:root-qerror}) that partially reflects propagation effects~\cite{ioannidis1991}, and (c) conservative policies (join-only, blending, fallback) intended to make a future in-planner integration low-risk.
We further specify a safe-injection layer (Section~\ref{sec:safe-injection}) to prevent semantic inconsistencies and reduce plan instability when integrating corrections into a cost model.

\subsection{Minimal end-to-end study design (future work)}
\label{sec:future-e2e}

To directly address the end-to-end question without over-claiming, we outline a minimal study design that we plan to implement as future work:
\begin{itemize}
\item \textbf{Shadow-mode planning:} for each query, generate plans under (i) native CE and (ii) TiCard-corrected CE, log plan choices and estimated costs, but execute only the production plan initially.
\item \textbf{Safety-first rollout:} enable join-only correction and bounded blending first; expand scope only when plan stability and regressions are acceptable on shadow traffic.
\item \textbf{Attribution controls:} keep the enumerator and cost model unchanged; vary only the cardinality interface to isolate the impact of CE corrections.
\item \textbf{Metrics:} plan changes (join order / operator choices), compile-time overhead, and execution-time distributions (median and tail latency), plus rollback triggers.
\end{itemize}

In TiDB, this study can be implemented with minimal operational risk by leveraging SQL Plan Management (SPM) to \emph{bind} and \emph{unbind} plans dynamically~\cite{tidb_spm}.
Concretely, once TiCard discovers a consistently better plan for a query (e.g., under join-only correction and bounded blending), the system can asynchronously bind that plan via SPM, monitor runtime/variance, and roll back by unbinding if regressions are detected.
This mechanism provides a practical pathway to staged deployment without requiring immediate, invasive changes to the optimizer's plan enumeration logic.

This design complements our offline evidence and aligns with deployability goals: incremental adoption with clear observability and reversibility.

\subsection{Safe injection: consistency constraints and stability guards}
\label{sec:safe-injection}

A common concern with per-operator learning is that naively injecting independent predictions into a cost model can increase plan instability or violate operator semantics (e.g., a filter producing more rows than its input).
TiCard's design therefore assumes a \emph{safe-injection layer} between model inference and the optimizer's cost model.
This layer is lightweight (pure post-processing on estimated cardinalities) and can be applied before any planner-in-the-loop rollout.

\paragraph{What can go wrong without guards.}
Even if each node's prediction reduces marginal Q-error, a plan can still be harmed if (i) a few nodes receive extreme corrections, (ii) corrected cardinalities become inconsistent with operator semantics, or (iii) changes in a small subset of nodes flip join-order decisions.
These are not model failures per se; they are integration risks.

\paragraph{Guard families.}
We recommend four guard families that together address the above risks while preserving TiCard's ``augment, not replace'' stance:
\begin{itemize}
\item \textbf{Semantic bounds (local):} enforce operator-level invariants where they are unambiguous (Table~\ref{tab:constraints}).
\item \textbf{Bounded correction (global):} clamp correction factors to a safe range (e.g., \(\exp(\hat{y}) \in [c_{\min}, c_{\max}]\)) or to a validation-calibrated quantile band, to prevent rare extreme outputs from dominating the cost model.
\item \textbf{Scope gating (structural):} apply corrections only to high-leverage operator classes first (e.g., joins) and fall back to native estimates elsewhere; expand scope only after shadow-mode evidence is satisfactory.
\item \textbf{Plan-change gating (decision-level):} when corrected CE changes the chosen plan, require a minimum predicted benefit threshold and allow rollback (e.g., via TiDB SPM binding/unbinding)~\cite{tidb_spm}.
\end{itemize}

\begin{table}[t]
\centering
\caption{Examples of semantics-aware constraints for safe injection (enforced on corrected cardinalities \(\widehat{C}\)).}
\label{tab:constraints}
\small
\setlength{\tabcolsep}{3.0pt}
\begin{tabularx}{\columnwidth}{@{}p{0.36\columnwidth}L@{}}
\toprule
\textbf{Operator class} & \textbf{Safe constraint (examples)} \\
\midrule
Filter / Selection & \(\widehat{C}_{out} \leftarrow \min(\widehat{C}_{out}, \widehat{C}_{in})\) \\
Limit & \(\widehat{C}_{out} \leftarrow \min(\widehat{C}_{out}, \text{limit}, \widehat{C}_{in})\) \\
Projection & \(\widehat{C}_{out} \leftarrow \widehat{C}_{in}\) (or \(\le\) if system allows dedup/other effects) \\
Aggregation (group-by) & \(\widehat{C}_{out} \leftarrow \min(\widehat{C}_{out}, \widehat{C}_{in})\) \\
Inner join & No general monotone bound holds; rely on bounded correction + join-only gating \\
Left outer join & \(\widehat{C}_{out} \leftarrow \max(\widehat{C}_{out}, \widehat{C}_{left})\) \\
\bottomrule
\end{tabularx}
\end{table}

\paragraph{Implementation sketch.}
In a TiDB integration, these guards can be implemented as a deterministic post-processing pass over the plan tree: apply bounded correction to each node, then enforce local semantic constraints where applicable (e.g., top-down for monotone operators such as filters/limits, and operator-specific rules for outer joins).
This ensures corrected cardinalities are (a) non-negative, (b) within a validated range relative to the native estimate, and (c) consistent with basic operator semantics before the cost model consumes them.

\paragraph{What we claim.}
Our offline results evaluate the \emph{prediction quality} of residual correction and conservative policies such as join-only and blending.
The safe-injection constraints above are proposed integration mechanisms to prevent pathological planner behavior; they do not require new training data and are compatible with shadow-mode rollout.

\section{Related Work}
\label{sec:related}

\subsection{Traditional cardinality estimation}

Histograms and independence-based estimators are the dominant classical approach~\cite{ioannidis2003,poosala1996,bruno2001}, but cannot capture cross-table correlations.
Sampling and sketching methods improve robustness at the cost of runtime overhead or limited query expressiveness~\cite{haas1995,chaudhuri1998,cormode2005}.

\subsection{Learned and hybrid estimation}

Query-driven learned estimators map query features to cardinalities (e.g., MSCN)~\cite{kipf2019learned}, while data-driven approaches learn joint distributions (e.g., NeuroCard/DeepDB/BayesCard)~\cite{yang2020neurocard,hilprecht2020deepdb,kipf2019bayescard}.
Hybrid approaches and correction methods aim to combine learned models with native estimators~\cite{wang2021face,wu2021uae,negi2023}.
LiteCard learns many lightweight local models keyed by repeated patterns with hierarchical fallback~\cite{yi2025biggerbreadboxefficientcardinality}.
TiCard shares a correction-first philosophy but emphasizes EXPLAIN-only features and deployability-oriented model choices (GBR vs.\ in-context TabPFN).

\paragraph{Transfer and benchmarks.}
Recent work studies transfer and pretraining for CE (e.g., PRICE)~\cite{zeng2024price} and highlights evaluation challenges and generalization gaps across schemas/workloads (e.g., CardBench)~\cite{chronis2024cardbench}.
These directions are highly relevant to deployability, but they typically require richer query/predicate encodings or access to database contents/statistics beyond the EXPLAIN-only interface we target.
As a result, we treat them as complementary baselines and discuss integration opportunities rather than claiming direct empirical comparability in our EXPLAIN-only, low-intrusion scope.

\paragraph{Deployable hybrid/data-driven models.}
Several recent estimators emphasize end-to-end integration and practical overhead, including FactorJoin (hybrid PGM framework for join queries)~\cite{wu2023factorjoin} and workload- or data-aware SPN-based approaches~\cite{liu2025qspn}.
Other work explores decoupled predicate modulation for joins~\cite{zhang2025distjoin} or learning under imperfect workloads~\cite{wu2025grasp}.
These methods are highly relevant comparators from a deployability perspective; however, they typically require access to base-table data/statistics and query predicates, and their reported end-to-end benefits arise from system-specific integration into the planner and/or execution feedback loops.
TiCard targets a different deployment surface: EXPLAIN-only features and correction policies that can be inserted conservatively as an augmentation layer, with a roadmap for planner-in-the-loop evaluation.

\subsection{ML for query optimization}

Learned optimizers and cost models apply ML to plan selection or runtime prediction~\cite{marcus2021bao,yang2022balsa,pathak2025}.
TiCard is complementary: it targets a narrow, deployable interface (cardinality corrections) that can be combined with other optimizer components.

\section{Conclusion}
\label{sec:conclusion}

TiCard reframes learned CE for deployability: instead of replacing the optimizer, it learns residual corrections over native estimates using EXPLAIN-only features.
In an offline, operator-level study on TiDB with TPC-H and JOB under a low-trace regime, TiCard reduces tail errors dramatically (P90 from 312.85 to 13.69 with GBR; P99 from 37,974.37 to 3,416.50 with TabPFN) while enabling conservative integration policies such as join-only correction.
We provide an integration roadmap to connect this offline evidence to online optimizer use, and view TiCard as a practical AI4DB component for incremental adoption in production databases.

\section*{Acknowledgements}
This work was supported by PingCAP, the company behind TiDB.

\section*{Declarations}
\textbf{Funding} Not applicable.\\
\textbf{Conflict of interest/Competing interests} The author is employed by PingCAP.\\
\textbf{Ethics approval and consent to participate} Not applicable.\\
\textbf{Consent for publication} Not applicable.\\
\textbf{Data availability} Query plans used in the offline evaluation are cached under \texttt{query\_plans/} in the accompanying repository. Large raw datasets (TPC-H tables, IMDB) follow their original licenses and are not redistributed.\\
\textbf{Materials availability} Not applicable.\\
\textbf{Code availability} Source code and scripts are available at \url{https://github.com/Icemap/TiCard}.\\
\textbf{Author contribution} Qizhi Wang: conceptualization, implementation, evaluation, and writing.

\bibliographystyle{unsrt}
\bibliography{references}

\end{document}